\pdfoutput=1
\documentclass[11pt]{article}

\usepackage{naacl2021}

\usepackage{times}
\usepackage{latexsym}

\usepackage[T1]{fontenc}
\usepackage[utf8]{inputenc}

\usepackage{microtype}
\usepackage{mathtools}
\usepackage{multirow}
\usepackage{tabularx}
\usepackage{booktabs}
\usepackage{xspace}
\usepackage{xcolor,colortbl}
\usepackage{arydshln}
\usepackage{subcaption}
\usepackage{paralist}
\usepackage{amsmath,amsfonts,amssymb}
\usepackage{relsize}

\usepackage[]{quoting}
\usepackage[warn]{textcomp}
\usepackage{silence}
\usepackage{ascii}
\usepackage[russian, main=english]{babel}

\usepackage{cleveref}
\crefname{section}{§}{§§}

\usepackage{geometry}

\usepackage{algorithm}
\usepackage{algorithmic}

\newcommand{\en}{\textsc{en}\xspace}
\newcommand{\de}{\textsc{de}\xspace}
\newcommand{\ita}{\textsc{it}\xspace} % \it is already defined
\newcommand{\fr}{\textsc{fr}\xspace}
\newcommand{\hr}{\textsc{hr}\xspace}
\newcommand{\ru}{\textsc{ru}\xspace}
\newcommand{\fin}{\textsc{fi}\xspace} % \fi is already defined
\newcommand{\tr}{\textsc{tr}\xspace}
\newcommand{\es}{\textsc{es}\xspace}
\newcommand{\zh}{\textsc{zh}\xspace}

\newcommand{\procb}{\textsc{Proc-B}\xspace}
\newcommand{\vecmap}{\textsc{VecMap}\xspace}
\newcommand{\muse}{\textsc{Muse}\xspace}
\newcommand{\joint}{\textsc{JA}\xspace}

\newcommand{\lone}{$\ell_{1}$}
\newcommand{\ltwo}{$\ell_{2}$}

\DeclareMathOperator*{\argmin}{argmin} 
\DeclareMathOperator{\trace}{tr}
\DeclareMathOperator{\sgn}{sgn}

\newcolumntype{P}[1]{>{\centering\arraybackslash}p{#1}}
\newcolumntype{Y}{>{\centering\arraybackslash}X}

\usepackage{microtype}

\title{Cross-Lingual Word Embedding Refinement by $\boldsymbol\ell_{1}$ Norm Optimisation}

\author{Xutan Peng \hspace{4mm}  Chenghua Lin\thanks{~~Chenghua Lin is the corresponding author.}~ \hspace{4mm}  Mark Stevenson\\
  Department of Computer Science, The University of Sheffield, UK \\
  \texttt{\{x.peng, c.lin, mark.stevenson\}@shef.ac.uk}
}

\begin{document}

\maketitle

\begin{abstract}
Cross-Lingual Word Embeddings (CLWEs) encode words from two or more languages in a shared high-dimensional space in which vectors representing words with similar meaning (regardless of language) are closely located. Existing methods for building  high-quality CLWEs learn mappings that minimise the \ltwo\ norm loss function.
However, this optimisation objective has been demonstrated to be sensitive to outliers.
Based on the more robust Manhattan norm (aka. \lone\ norm) goodness-of-fit criterion, this paper proposes a simple post-processing step to improve CLWEs. 
An advantage of this approach is that it is fully agnostic to the training process of the original CLWEs and can therefore be applied widely. 
Extensive experiments are performed involving ten diverse languages and embeddings trained on different corpora. 
Evaluation results based on bilingual lexicon induction and cross-lingual transfer for natural language inference tasks show that the \lone\ refinement substantially outperforms four state-of-the-art baselines in both supervised and unsupervised settings.
It is therefore recommended that this strategy be adopted as a standard for CLWE methods.
\end{abstract}

\section{Introduction}\label{sec:intro}

Cross-Lingual Word Embedding (CLWE) techniques have recently received significant attention as an effective means to support Natural Language Processing applications for low-resource languages, e.g., machine translation~\citep{artetxe2018unsupervised} and transfer learning~\citep{HistSumm-2021}.

The most successful CLWE models are the so-called \textit{projection-based} methods, which learn mappings between monolingual word vectors with very little, or even zero, cross-lingual supervision~\citep{muse,vecmap,xling}.
Mainstream projection-based CLWE models typically identify orthogonal mappings by minimising the topological dissimilarity between source and target embeddings based on \ltwo\ loss (aka. Frobenius loss or squared error)~\cite{xling,Ruder_survey}. 
This learning strategy has two  advantages. First, adding the orthogonality constraint to the mapping function has been demonstrated to significantly enhance the quality of CLWEs~\citep{xing-2015}.
Second, the existence of a closed-form solution to the \ltwo\ optima~\citep{l2_svd} greatly simplifies the computation required~\citep{artetxe-2016,proof_orthogonal}.

Despite its popularity, work in various application domains has noted that \ltwo\ loss is not robust to noise and outliers. It is widely known in computer vision that \ltwo-loss-based solutions can severely exaggerate noise, leading to inaccurate estimates~\citep{cv_l2_svd,cv_l1_re}. In data mining, Principal Component Analysis (PCA) using \ltwo\ loss has been shown to be sensitive to the presence of outliers in the input data, degrading the quality of the feature space produced~\citep{dm_l1_pca}.
Previous studies have demonstrated that the processes used to construct monolingual and cross-lingual embeddings may introduce noise (e.g. via reconstruction error~\citep{reconstruction-error} and structural variance~\citep{Ruder_survey}), making the presence of outliers more likely. Empirical analysis of CLWEs also demonstrates that more distant word pairs (which are more likely to be outliers) have more influence on the behaviour of \ltwo\ loss than closer pairs. This raises the question of the appropriateness of \ltwo\ loss functions for CLWEs.

Compared to the conventional \ltwo\ loss, \lone\ loss (aka. Manhattan distance) has been mathematically demonstrated to be less affected by outliers~\citep{l2_bad} and empirically proven useful in computer vision and data mining~\citep{cv_l2_svd,cv_l1_re,dm_l1_pca}.
Motivated by this insight, our paper proposes a simple yet effective post-processing technique to improve the quality of CLWEs: adjust the alignment of \textit{any} cross-lingual vector space to minimise the \lone\ loss without violating the orthogonality constraint.
Specifically, given existing CLWEs, we bidirectionally retrieve bilingual vectors and optimise their Manhattan distance using a numerical solver.
The approach can be applied to any CLWEs, making the post-hoc refinement technique generic and applicable to a wide range of scenarios.
We believe this to be the first application of \lone\ loss to the CLWE problem.

To demonstrate the effectiveness of our method, we select four state-of-the-art baselines and conduct comprehensive evaluations in both supervised and unsupervised settings.
Our experiments involve ten languages from diverse branches/families and embeddings trained on corpora of different domains.
In addition to the standard Bilingual Lexicon Induction (BLI) benchmark, we also investigate a downstream task, namely cross-lingual transfer for Natural Language Inference (NLI). 
In all setups tested, our algorithm significantly improves the performance of strong baselines. 
Finally, we provide an intuitive visualisation illustrating why \lone\ loss is more robust than it \ltwo\ counterpart when refining CLWEs (see Fig.~\ref{fig:gap}).  
Our code is available at \url{https://github.com/Pzoom522/L1-Refinement}.

Our contribution is three-fold: 
(1) we propose a robust refinement technique based on the \lone\ norm training objective, which can effectively enhance CLWEs;
(2) our approach is generic and can be directly coupled with both supervised and unsupervised CLWE models; 
(3) our \lone\ refinement algorithm achieves state-of-the-art performance for both BLI and cross-lingual transfer for NLI tasks.

\section{Related Work}\label{sec:rw}

\paragraph{CLWE methods.}
One approach to generating CLWEs is to train shared semantic representations using multilingual texts aligned at sentence or document level~\citep{vulic-2016-sentence,upadhyay-2016-sentence}.
Although this research direction has been well studied, the parallel setup requirement for model training is expensive, and hence impractical for low-resource languages.

Recent years have seen an increase in interest in projection-based methods, which train CLWEs by finding mappings between pretrained word vectors of different languages~\cite{mikolov_linear_mapping,muse,peng2020revisiting}. 
Since the input embeddings can be generated independently using monolingual corpora only, projection-based methods reduce the supervision required for training and offer a viable solution for low-resource scenarios.

\citet{xing-2015} showed that the precision of the learned CLWEs can be improved by constraining the mapping function to be orthogonal, which is formalised as the so-called \ltwo\ Orthogonal Procrustes Analysis (OPA):
\begin{equation}\label{eq:l2_opp}
    \argmin_{\mathbf{M} \in \mathcal{O}} \Vert \mathbf{A} \mathbf{M} - \mathbf{B} \Vert_{2},
\end{equation}
where $\mathbf{M}$ is the CLWE mapping, $\mathcal{O}$ denotes the orthogonal manifold (aka. the Stiefel manifold~\citep{manifold}), and $\mathbf{A}$ and $\mathbf{B}$ are matrices composed using vectors from source and target embedding spaces.

While \citet{xing-2015} exploited an approximate and relatively slow gradient-based solver, more recent approaches such as \citet{artetxe-2016} and \citet{proof_orthogonal} introduced an exact closed-form solution for Eq.~\eqref{eq:l2_opp}.
Originally proposed by \citet{l2_svd}, it utilises Singular Value Decomposition (SVD): 
\begin{equation}\label{eq:opp_solution}
    \mathbf{M}^{\star} = \mathbf{U} \mathbf{V}^\intercal, \text{with } \mathbf{U}\Sigma \mathbf{V}^\intercal = \mathrm{SVD}(\mathbf{A}^\intercal \mathbf{B}),
\end{equation}
where $\mathbf{M}^{\star}$ denotes the \ltwo-optimal mapping matrix. 
The efficiency and effectiveness of Eq.~\eqref{eq:opp_solution} have led to its application within many other approaches, e.g., \citet{dlv}, \citet{rcsls} and \citet{xling}.
In particular, \procb~\citep{xling}, a supervised CLWE framework that simply applies multiple iterations of \ltwo\ OPA, has been demonstrated to produce very competitive performance on various benchmark tasks including BLI as well as cross-lingual transfer for NLI and information retrieval.

While the aforementioned approaches still require some weak supervision (i.e., seed dictionaries), there have also been some successful attempts to train CLWEs in a completely unsupervised fashion.
For instance, \citet{muse} proposed a system called \muse, which bootstraps CLWEs without any bilingual signal through adversarial learning. 
\vecmap~\citep{vecmap} applied a self-learning strategy to iteratively compute the optimal mapping and then retrieve bilingual dictionary.
Comparing \muse and \vecmap, the latter tends to be more robust as its similarity-matrix-based heuristic initialisation is more stable in most cases~\citep{xling,Ruder_survey}.
Very recently, some studies bootstrapped unsupervised CLWEs by jointly training word embeddings on concatenated corpora of different languages and achieved good performance \cite{joint}.

\paragraph{The \ltwo\ refinement algorithm.} 
CLWE models often apply \ltwo\ refinement, a post-processing step shown to improve the quality of the initial alignment (see \citet{Ruder_survey} for survey). 
Given existing CLWEs $\{\mathbf{X}_{\mathtt{L_A}}$, $\mathbf{X}_{\mathtt{L_B}}\}$ for languages $\mathtt{L_A}$ and $\mathtt{L_B}$, bidirectionally one can use approaches such as the classic nearest-neighbour algorithm, the inverted softmax~\citep{proof_orthogonal} and the cross-domain similarity local scaling (CSLS)~\citep{muse} to retrieve two bilingual dictionaries $D_{\mathtt{L_A} \mapsto \mathtt{L_B}}$ and $D_{\mathtt{L_B} \mapsto \mathtt{L_A}}$.
Note that word pairs in $D_{\mathtt{L_A} \mapsto \mathtt{L_B}} \cap D_{\mathtt{L_B} \mapsto \mathtt{L_A}}$ are highly reliable, as they form ``mutual translations''.
Next, one can compose bilingual embedding matrices $\mathbf{A}$ and $\mathbf{B}$ by aligning word vectors (rows) using the above word pairs.
Finally, a new orthogonal mapping is learned to fit $\mathbf{A}$ and $\mathbf{B}$ based on least-square regressions, i.e., perform \ltwo\ OPA described in Eq.~\eqref{eq:l2_opp}.

Early applications of \ltwo\ refinement applied a \textit{single} iteration, e.g. \cite{vulic-2016-sentence}. Due to the wide adoption of the closed-form \ltwo\ OPA solution (cf. Eq.~\eqref{eq:opp_solution}), recent methods perform multiple iterations. 
The iterative \ltwo\ refinement strategy is an important component of approaches that bootstrap from small or null training lexicons~\citep{vecmap}. However, a single step of refinement is often sufficient to create suitable CLWEs~\citep{muse,xling}.

\section{Methodology}\label{sec:method}
A common characteristic of CLWE methods that apply the orthogonality constraint is that they optimise using \ltwo\ loss (see \cref{sec:rw}).
However, outliers have disproportionate influence in \ltwo\ since the penalty increases quadratically and this can be particularly  problematic with noisy data since the solution can ``shift'' towards them~\citep{l2_bad}. 
The noise and outliers present in real-world word embeddings may affect the performance of \ltwo-loss-based CLWEs.

The \lone\ norm cost function is more robust than \ltwo\ loss as it is less affected by outliers~\citep{l2_bad}.
Therefore, we propose a refinement algorithm for improving the quality of CLWEs based on \lone\ loss. 
This novel method, which we refer to as \lone\ refinement, is generic and can be applied post-hoc to improve the output of existing CLWE models.
To our knowledge, the use of alternatives to \ltwo-loss-based optimisation has never been explored by the CLWE community.

To begin with, analogous to \ltwo\ OPA (cf. Eq.~\eqref{eq:l2_opp}), \lone\ OPA can be formally defined and rewritten as
\begin{align}\label{eq:l1_opp}
      & \argmin_{ {\mathbf{M} \in \mathcal{O}}} \Vert \mathbf{A} \mathbf{M}  -  \mathbf{B} \Vert_{1} \nonumber \\
    =  & \argmin _{\mathbf{M}  \in \mathcal{O}} \trace [ (  \mathbf{A} \mathbf{M}  -  \mathbf{B} )^\intercal \sgn(\mathbf{A} \mathbf{M}  -  \mathbf{B}) ],
\end{align}
where $\trace(\cdot)$ returns the matrix trace,  $\sgn(\cdot)$ is the signum function, and $\in \mathcal{O}$ denotes that $M$ is subject to the orthogonal constraint.
Compared to \ltwo\ OPA which has a closed-form solution, solving Eq.~\eqref{eq:l1_opp} is much more challenging due to the discontinuity of $\sgn(\cdot)$. 
This issue can be addressed by replacing $\sgn(\cdot)$ with $\tanh (\alpha(\cdot))$, a smoothing function parameterised by $\alpha$, such that
\begin{equation}\label{eq:smoothed_l1_opp}
    \argmin_{\mathbf{M} \in \mathcal{O}} \trace [ (\mathbf{ A} \mathbf{ M} \! - \! \mathbf{ B})^\intercal \tanh (\alpha(\mathbf{ A} \mathbf{M} \! - \! \mathbf{ B})) ].
\end{equation}
Larger values for $\alpha$ lead to closer approximations to $\sgn(\cdot)$ but reduce the smoothing effect. This approach has been used in many applications, such as the activation function of long short-term memory networks~\citep{lstm}.

However, in practice, we find that Eq.~\eqref{eq:smoothed_l1_opp} remains unsolvable in our case with standard gradient-based frameworks for two reasons. 
First, $\alpha$ has to be sufficiently large in order to achieve a good approximation of $\sgn(\cdot)$. 
Otherwise, relatively small residuals will be down-weighted during fitting and the objective will become biased towards outliers, just similar to \ltwo\ loss. 
However, satisfying this requirement (i.e., large $\alpha$) will lead to the activation function $\tanh (\alpha(\cdot))$ becoming easily \textit{saturated}, resulting in an optimisation process that becomes trapped during the early stages. In other words, the optimisation can only reach an unsatisfactory local optimum. Second, the orthogonality constraint (i.e., $\mathbf{M} \in \mathcal{O}$) also makes the optimisation more problematic for these methods.

We address these challenges by adopting the approaches proposed by \citet{l1}. This method explicitly encourages the solver to only explore the desired manifold $\mathcal{O}$ thereby reducing the \lone\ solver's search space and difficulty of the optimisation problem. We begin by calculating the gradient $\nabla$ w.r.t. the objective in Eq.~\eqref{eq:smoothed_l1_opp} through matrix differentiation: 
\begin{equation}\label{eq:define_nabla}
    \nabla = \mathbf{A} ^\intercal(\tanh(\mathbf{Z}) + \mathbf{Z} \odot \cosh^{-2}(\mathbf{Z})),
\end{equation}
where $\mathbf{Z}$$=$$\alpha (\mathbf{A} \mathbf{M}$$-$$\mathbf{B})$ and $\odot$ is the Hadamard product. 
Next, to find the steepest descent direction while ensuring that any $\mathbf{M}$ produced is orthogonal, we project $\nabla$ onto $\mathcal{O}$, yielding\footnote{See \citet{manifold} for derivation details.}
\newcommand{\Scale}[1]{\scalebox{1.0}{#1}}
\begin{equation}\label{eq:projected_grad}
    \pi_\mathcal{O}(\nabla) \!\! := \!\! \frac{\Scale 1}{\Scale 2} \mathbf{M} (\mathbf{M} ^\intercal \nabla \! - \! \nabla ^\intercal\mathbf{M}) + (\mathbf{I} \! - \! \mathbf{M} \mathbf{M} ^\intercal)\nabla.
\end{equation}
Here $\mathbf{I}$ is an identity matrix with the shape of $\mathbf{M}$.
With Eq.~\eqref{eq:projected_grad} defining the optimisation flow, our \lone\ loss minimisation problem reduces to an integration problem, as
\begin{equation}\label{eq:ode}
    \mathbf{M^\star} = \mathbf{M_0} + \mathsmaller{\int} -\pi_\mathcal{O}(\nabla)\,\mathrm {d} t,
\end{equation}
where $\mathbf{M_0}$ is a proper initial solution of Eq.~\eqref{eq:l1_opp} (e.g., \ltwo-optimal mapping obtained via Eq.~\eqref{eq:opp_solution}).

Empirically, unlike the aforementioned standard gradient-based methods, by following the established policy of Eq.~\eqref{eq:projected_grad}, the optimisation process of Eq.~\eqref{eq:ode} will not violate the orthogonality restriction or get trapped during early stages.
However, this \lone\ OPA solver requires extremely small step size to generate reliable solutions~\citep{l1}, making it computationally expensive\footnote{It takes averagely 3 hours and up to 12 hours to perform Eq.~\eqref{eq:ode} on an Intel Core i9-9900K CPU. In comparison, the time required to solve Eq.~\eqref{eq:opp_solution} in each training loop is less than 1 second and the iterative \ltwo-norm-based training takes 1 to 5 hours in total.}.
Therefore, it is impractical to perform \lone\ refinement in an iterative fashion like \ltwo\ refinement without significant computational resources.

Previous work has demonstrated that applying the \lone-loss-based algorithms from a good initial state can speed up the optimisation.
For instance, \citet{dm_l1_pca} found that feature spaces created by \ltwo\ PCA were severely affected by noise. Replacing the cost function with \lone\ loss significantly reduced this problem, but required expensive linear programming. To reduce the convergence time, \citet{fast_l1_pca} exploited the first principal component from the \ltwo\ solution as an initial guess. Similarly, when reconstructing corrupted pixel matrices, \ltwo-loss-based results are far from satisfactory; using \lone\ norm estimators can improve the quality, but are too slow to handle large-scale datasets~\cite{cv_l2_svd}. However, taking the \ltwo\ optima as the starting point allowed less biased reconstructions to be learned in an acceptable time \cite{cv_l1_re}.

\begin{algorithm}[!t]\small
\caption{\lone\ refinement}
\begin{algorithmic}[1]
\REQUIRE CLWEs $\{ \mathbf{X}_{\mathtt{L_A}}, \mathbf{X}_{\mathtt{L_B}} \}$
\ENSURE updated CLWEs $\{ \mathbf{X}_{\mathtt{L_A}}\mathbf{M^\star}, \mathbf{X}_{\mathtt{L_B}} \}$

\STATE $D_{\mathtt{L_A} \mapsto \mathtt{L_B}} \leftarrow$ build dict via $\mathbf{X}_{\mathtt{L_A}}$ and $\mathbf{X}_{\mathtt{L_B}}$ 
\STATE $D_{\mathtt{L_B} \mapsto \mathtt{L_A}} \leftarrow$ build dict via $\mathbf{X}_{\mathtt{L_B}}$ and $\mathbf{X}_{\mathtt{L_A}}$  
\STATE $D \leftarrow D_{\mathtt{L_A} \mapsto \mathtt{L_B}} \cap D_{\mathtt{L_B} \mapsto \mathtt{L_A}}$
\STATE $\mathbf{A}, \mathbf{B} \leftarrow $ looks up for $D$ in $\mathbf{X}_{\mathtt{L_A}}, \mathbf{X}_{\mathtt{L_B}}$
\STATE perform integration to solve Eq.~\eqref{eq:ode} for $\mathbf{M^\star}$, with initial value $\mathbf{M_0} \leftarrow \mathbf{I}$, until stopping criteria are met
\end{algorithmic}\label{algo:l1}
\end{algorithm}

Inspired by these works, we make use of \lone\ refinement to carry out post-hoc enhancement of existing CLWEs.
Our full pipeline is described in Algorithm~\ref{algo:l1} (see \cref{ssec:solver} for implemented configurations).
In common with \ltwo\ refinement (cf. \cref{sec:rw}), steps 1-4 bootstrap a synthetic dictionary $D$ and compose bilingual word vector matrices $\mathbf{A}$ and $\mathbf{B}$ which have reliable row-wise correspondence.
Taking them as the starting state, in step 5 an identity matrix naturally serves as our initial solution $\mathbf{M_0}$.

During the execution of Eq.~\eqref{eq:ode}, we record \lone\ loss per iteration and see if \textit{either} of the following two stopping criteria have been satisfied: (1) the updated \lone\ loss exceeds that of the previous iteration; (2) on-the-fly $\mathbf{M}$ has non-negligibly departed from the orthogonal manifold, which can be indicated by the maximum value of the disparity matrix as 
\begin{equation}\label{eq:err_threshold}
    \mathrm{max} (|\mathbf{M} ^\intercal \mathbf{M} - \mathbf{I}|) > \epsilon,
\end{equation}
where $\epsilon$ is a sufficiently small threshold.
The resulting $\mathbf{M^\star}$ can be used to adjust the word vectors of $\mathtt{L_A}$ and output refined CLWEs.

A significant advantage of our algorithm is its generality: it is fully independent of the method used for creating the original CLWEs and can therefore be used to enhance a wide range of models, both in supervised and unsupervised settings.

\section{Experimental Setup}

\subsection{Datasets} 

In order to demonstrate the generality of our proposed method, we conduct experiments using two groups of monolingual word embeddings trained on very different corpora:

\noindent\textbf{Wiki-Embs} \citep{157}: embeddings developed using Wikipedia dumps for a range of ten diverse languages: two Germanic (English{\scriptsize$|$\en}, German{\scriptsize$|$\de}), two Slavic (Croatian{\scriptsize$|$\hr}, Russian{\scriptsize$|$\ru}), three Romance  (French{\scriptsize$|$\fr}, Italian{\scriptsize$|$\ita}, Spanish{\scriptsize$|$\es}) and three non-Indo-European (Finnish{\scriptsize$|$\fin} from the Uralic family, Turkish{\scriptsize$|$\tr} from the Turkic family and Chinese{\scriptsize$|$\zh} from the Sino-Tibetan family).

\noindent\textbf{News-Embs} \citep{vecmap}: embeddings trained on a multilingual News text collection, i.e., the WaCKy Crawl of $\{\en, \de, \ita\}$, the Common Crawl of \fin, and the WMT News Crawl of \es. 

News-Embs are considered to be more challenging for building good quality CLWEs due to the heterogeneous nature of the data, while a considerable portion of the multilingual training corpora for Wiki-Embs are roughly parallel. Following previous studies~\citep{muse,vecmap,density,xling}, only the first 200K vocabulary entries are preserved.

\subsection{Baselines} 

\citet{xling} provided a systematic evaluation for projection-based CLWE models, demonstrating that three methods (i.e., \muse, \vecmap, and \procb) achieve the most competitive performance. A recent algorithm (\joint) by \citet{joint}  also reported state-of-the-art results. 
For comprehensive comparison, we therefore use all these four methods as the main baselines for both supervised and unsupervised settings:

\noindent\textbf{\muse}~\citep{muse}: an \textit{unsupervised} CLWE model based on adversarial learning and iterative \ltwo\ refinement;

\noindent\textbf{\vecmap}~\citep{vecmap}: a robust \textit{unsupervised} framework using a self-learning strategy;

\noindent\textbf{\procb}~\citep{xling}: a simple but effective \textit{supervised} approach to creating CLWEs;

\noindent \textbf{\joint-\muse} and \textbf{\joint-RCSLS}~\cite{joint}: 
a recently proposed Joint-Align (\textbf{JA}) Framework, which first initialises CLWEs using joint embedding training, followed by vocabularies reallocation. It then utilises off-the-shelf CLWE methods to improve the alignment in both \textit{unsupervised} (\textbf{\joint-\muse}) and \textit{supervised} (\textbf{\joint-RCSLS}) settings.

In the original implementations, \muse, \procb and \joint were only trained on Wiki-Embs while \vecmap additionally used News-Embs. 
Although all baselines reported performance for BLI, they used various versions of evaluation sets, hence previous results are not directly comparable with the ones reposted here.
More concretely, the testsets for \muse/\joint and \vecmap are two different batches of \en-centric dictionaries, while the testset for \procb also supports non-\en translations.

\subsection{Implementation Details of Algorithm~\ref{algo:l1}}\label{ssec:solver}

The CSLS scheme with a neighbourhood size of 10 (CSLS-10) is adopted to build synthetic dictionaries via the input CLWEs.  
A variable-coefficient ordinary differential equation (VODE) solver\footnote{\url{http://www.netlib.org/ode/vode.f}} was implemented for the system described in Eq.~\eqref{eq:ode}. Suggested by~\citet{l1}, we set the maximum order at 15, the smoothness coefficient $\alpha$ in Eq.~\eqref{eq:define_nabla} at 1e8, the threshold $\epsilon$ in Eq.~\eqref{eq:err_threshold} at 1e-5, and performed the integration with a fixed time interval of 1e-6.
An early-stopping design was adopted to ensure computation completed in a reasonable time: in addition to the two default stopping criteria in \cref{sec:method}, integration is terminated if $\int\mathrm{d} t$ reaches 5e-3 ($\mathrm{d} t$ is the differentiation term in Eq.~\eqref{eq:ode}).

In terms of the tolerance of the VODE solver, we set the absolute tolerance at 1e-7 and the relative tolerance at 1e-5, following the established approach of \citet{vode-tor}. These tolerance settings show good generality empirically and were used for all tested language pairs, datasets, and models in our experiments.

\begin{table*}[!t]\small
\centering
\begin{subtable}{0.51\textwidth}
\begin{tabular}{{p{2.3cm}P{1.1cm}P{1.1cm}P{1.1cm}P{1.1cm}P{1.1cm}}}
\toprule 
& \en--\de & \en--\es & \en--\fr & \en--\ru & \en--\zh\\ \midrule
\muse\textsuperscript{\ETX}   &  74.0 &  81.7  &  82.3   & 44.0 & 32.5 \\
\muse-\ltwo & 74.0 &  82.1  &  82.6  & \textcolor{white}{*}43.8* & \textcolor{white}{*}31.9*\\ \hdashline
\muse-\lone & 75.2 &  82.6  &  82.9  & \textcolor{white}{*}45.6* &\textcolor{white}{*}33.8*\\
\midrule
\joint-\muse\textsuperscript{\EOT}   &  74.2 & 81.4 & 82.8 & 45.0 & 36.1  \\
\joint-\muse-\ltwo   & 74.1 & 81.6 & 82.7 & 45.1 & 36.2 \\ \hdashline
\joint-\muse-\lone   & \textbf{75.4} & 82.0  & \textbf{83.1} & 46.3 & \textbf{38.1}\\
\midrule
\vecmap\textsuperscript{\ENQ}   &  75.1 &  82.3  &  80.0   & 49.2 & \cellcolor{gray!20}00.0\\
\vecmap-\ltwo & 74.8 &  82.3  &  79.4  & 48.9 & \cellcolor{gray!20}00.0 \\ \hdashline
\vecmap-\lone & \textbf{75.4}&  \textbf{82.9}  &  80.2  & \textbf{49.9} & \cellcolor{gray!20}00.0 \\
\bottomrule
\end{tabular}
\caption{Wiki-Embs (setup of \citet{muse}).}
\label{tab:muse_dataset}
\end{subtable}
\hspace{\fill}
\begin{subtable}{0.43\textwidth}
\centering
\begin{tabular}{{p{2.3cm}P{1.1cm}P{1.1cm}P{1.1cm}P{1.1cm}}}
\toprule 
 & \en--\de & \en--\es & \en--\fin & \en--\ita \\ \midrule
\muse\textsuperscript{\ETX} & \cellcolor{gray!20}00.0 & \cellcolor{gray!20}07.1 & \cellcolor{gray!20}00.0 & \cellcolor{gray!20}09.1 \\
\muse-\ltwo & \cellcolor{gray!20}00.0 & \cellcolor{gray!20}00.0 & \cellcolor{gray!20}00.0 & \cellcolor{gray!20}00.0 \\ \hdashline
\muse-\lone & \cellcolor{gray!20}00.0 & \cellcolor{gray!20}00.0 & \cellcolor{gray!20}00.0 & \cellcolor{gray!20}00.0 \\
\midrule
\joint-\muse  &  47.9  & 48.4 & 33.0 & 37.2 \\ 
\joint-\muse-\ltwo   & 47.9 & 48.6 & 32.9 & 37.3 \\ \hdashline
\joint-\muse-\lone   & 48.8 & \textbf{49.7} & \textbf{35.2} & 37.7 \\
\midrule
\vecmap\textsuperscript{\ETX} & 48.2  & 48.1 &  32.6  &  37.3  \\ 
\vecmap-\ltwo & 48.1 & 47.9 &  32.9  & 37.1    \\ \hdashline
\vecmap-\lone & \textbf{49.0} & 48.9 &  34.4  & \textbf{37.8}   \\
\bottomrule
\end{tabular}
\caption{News-Embs (setup of \citet{vecmap}).}
\label{tab:vecmap_dataset}
\end{subtable}
\caption{ACC (\%) of unsupervised BLI.
(a) Rows marked with \ETX, \EOT and \ENQ are respectively from \citet{muse}, \citet{joint} and \citet{density}.
NB: for \en--$\{\ru, \zh\}$ we observed one failed run (ACC \textless 10.0\%), where we only record the average of successful scores with *.
(b) Rows marked with \ETX are from \citet{vecmap}.}
\label{tab:unsup_en}
\end{table*}
\begin{table*}[t]\small
\centering
\begin{subtable}{0.6\textwidth}
\begin{tabular}{{p{1.9cm}P{1.1cm}P{1.1cm}P{1.1cm}P{1.1cm}P{1.1cm}P{1.1cm}P{1.1cm}}}
\toprule
 & \en--\de & \en--\fin & \en--\fr & \en--\hr & \en--\ita & \en--\ru & \en--\tr \\ \midrule
\joint-RCSLS & 50.9 & 33.9 & 63.0 & 29.1 & 58.3 & 41.3 & 29.4 \\
\joint-RCSLS-\ltwo & 50.7 & 33.8 & 63.0 & 29.1 & 58.2 & 41.3 & 29.5 \\ \hdashline
\joint-RCSLS-\lone & 51.6 & 34.5 & 63.4 & 30.4 & 59.0 & 41.9 &30.2 \\ \midrule
\procb\textsuperscript{\ETX}& 52.1 & 36.0 & 63.3  & 29.6  & \textbf{60.5} & 41.9  & 30.1\\ 
\procb-\ltwo  & 51.8  & 34.4  & 63.1 & 28.2 & \textbf{60.5} & 39.8 & 28.0\\ \hdashline
\procb-\lone & \textbf{52.6} & \textbf{36.3} & \textbf{63.7} & \textbf{30.5}  & \textbf{60.5} & \textbf{42.3} & \textbf{30.9} \\ \bottomrule
\end{tabular}
\caption{Wiki-Embs (setup of \citet{xling}).}
\label{tab:procb_wiki}
\end{subtable}
\hspace{\fill}
\begin{subtable}{0.35\textwidth}
\centering
\begin{tabular}{{p{1.9cm}P{1.1cm}P{1.1cm}P{1.1cm}}}
\toprule 
 &\en--\de & \en--\fin & \en--\ita \\ \midrule
\joint-RCSLS & 46.8 & 42.0 & 37.4 \\
\joint-RCSLS-\ltwo & 46.9 & 42.2 & 37.5 \\ \hdashline
\joint-RCSLS-\lone & 48.3 & \textbf{44.6} & 39.0 \\
 \midrule
\procb & 47.5 & 41.4 & 37.3 \\
\procb-\ltwo & 47.1 & 41.7 & 37.4 \\ \hdashline
\procb-\lone & \textbf{52.6} & 43.3 & \textbf{41.1} \\
\bottomrule
\end{tabular}
\caption{News-Embs.}
\label{tab:procb_news}
\end{subtable}
\caption{MRR (\%) of supervised BLI.
Rows marked with \ETX are from the supplementary of \citet{xling}.}
\label{tab:procb_en}
\end{table*}

\section{Results}
We evaluate the effectiveness of the proposed \lone\ refinement technique on two benchmarks: Bilingual Lexicon Induction (BLI), the \textit{de facto} standard for measuring the quality of CLWEs, and a downstream natural language inference task based on  cross-lingual transfer. In addition to comparison against state-of-the-art CLWE models, we also report the performance of the single-iteration \ltwo\ refinement method which follows steps 1-4 of Algorithm~\ref{algo:l1} then minimises \ltwo\ loss in the final step.

To reduce randomness, we executed each model in each setup three times and the average accuracy (ACC, aka. precision at rank 1) is reported. Following \citet{xling}, by comparing scores achieved before and after \lone\ refinement, statistical significance is indicated via the $p$-value of two-tailed t-tests with Bonferroni correction~\citep{p-value} (note that $p$-values are not recorded for Tab.~\ref{tab:procb_news} given the small number of runs).

\subsection{Bilingual Lexicon Induction}\label{subsec:BLI}

\paragraph{Refining unsupervised baselines.}
Tab.~\ref{tab:muse_dataset} follows the main setup of \citet{muse}, who tested six language pairs using Wiki-Embs\footnote{Note that we are unable to report the result of English to Esperanto as the corresponding dictionary is missing, see \url{https://git.io/en-eo-dict-issue}.}. After \lone\ refinement, \muse-\lone, \joint-\muse-\lone, and \vecmap-\lone\ all significantly ($p < 0.01$) outperform their corresponding base algorithms, with an average 1.1\% performance gain over \muse, 1.1\% over \joint-\muse, and 0.5\% over \vecmap.   
To put these improvements in context, \citet{naacl19_weak} reported an improvement of 0.4\% for \vecmap on same dataset and language pairs.

Our method tends to work better on the more distant language pairs. 
For instance, for the distant pairs \en--$\{\ru, \zh\}$, the increments achieved by \muse-\lone\ are 1.6\% and 1.3\%, respectively;
whereas for the close pairs \en--$\{\de, \es, \fr\}$ the average gain is a maximum of 0.9\%. 
A similar trend can be observed for \joint-\muse-\lone\ and \vecmap-\lone. (As the \vecmap algorithm always collapses for  \en--\zh, no result is reported for this language pair).

Another set of experiments were conducted to evaluate the robustness of our algorithm following the main setup of \citet{vecmap}, who tested four language pairs based on the more homogeneous News-Embs. Tab.~\ref{tab:vecmap_dataset} shows that \joint-\muse-\lone\ and \vecmap-\lone\ consistently improves the original \vecmap with an average gain of 1.2\% and 1.0\% ($p$\textless0.01). Obtaining such substantial improvements over the state-of-the-art is nontrivial.   For example, even a very recent weakly supervised method by \citet{emnlp19_weak} is \textit{inferior} to \vecmap  by 1.0\% average ACC.
On the other hand, \muse fails to produce any analysable result as it always collapses on the more challenging News-Embs. 
Improvement with \lone\ refinement is also larger when language pairs are more distant, e.g., for \vecmap-\lone\ the ACC gain on \en-\fin is 1.8\%, more than double of the gain (0.7\%) on the close pairs \en--$\{\de,\ita\}$ (cf. Tab.~\ref{tab:muse_dataset} and above).

We also  conduct an ablation study by reporting the performance of \ltwo\ refinement scheme ($\{\muse, \joint\-\muse, \vecmap\}$-\ltwo).
This observation is in accordance with that of \citet{muse}, who reported that after performing \ltwo\ refinement in the first loop, applying further iterations only produces marginal precision gain, if any.

Overall, the \lone\ refinement consistently and significantly improve the CLWEs produced by base algorithms, regardless of the embeddings and setups used, thereby demonstrating the effectiveness and robustness of the proposed algorithm. 

\newcommand{\blkt}{\textcolor{white}{\scalebox{.65}{*}}}
\begin{table*}[!t]\small
\parbox{.58\linewidth}{
\centering
\begin{tabular}{{p{2.1cm}P{.95cm}P{.95cm}P{.95cm}P{.95cm}P{.95cm}P{.95cm}P{.95cm}}}
\toprule 
\textit{Unsupervised}& \de--\ita & \de--\tr & \fin--\hr & \fin--\ita & \hr--\ru & \ita--\fr  & \tr--\ita  \\ \midrule
ICP\textsuperscript{\ETX}& 44.7 & 21.5 & 20.8 & 26.3 & 30.9 & 62.9 & 24.3 \\  
GWA\textsuperscript{\ETX} & 44.0 & 10.1 & \cellcolor{gray!20}00.9 & 17.3 & \cellcolor{gray!20}00.1 & 65.5 & 14.2 \\ \midrule
\muse\textsuperscript{\ETX}& 49.6 & 23.7 & 22.8 & 32.7 & \cellcolor{gray!20}00.0 & 66.2 & 30.6 \\
\muse-\ltwo & 50.3 & 23.9 & 23.1 & 32.7 & 34.9 & 67.1 & \textcolor{white}{*}30.5* \\ \hdashline
\muse-\lone & 50.7 & 26.5 & 25.4 & 35.0 & 37.9 & 67.6 & \textcolor{white}{*}33.3* \\ \midrule
\joint-\muse & 50.9 & 25.6 & 23.4 & 34.9 & 36.9 & 68.3 & 34.7 \\
\joint-\muse-\ltwo & 50.9 & 25.5 & 23.4 & 34.7 & 36.9 & 68.4 & 34.7 \\ \hdashline
\joint-\muse-\lone & \textbf{51.5} & \textbf{28.4} & 26.1 & 36.0 & 37.6 & \textbf{68.7} & \textbf{36.1} \\ \midrule
\vecmap\textsuperscript{\ETX} & 49.3 & 25.3 & 28.0 & 35.5 & 37.6 & 66.7 & 33.2 \\ 
\vecmap-\ltwo & 48.8 & 25.7 & 28.5 & 35.8 & 38.4 & 67.0 & 33.5 \\ 
 \hdashline
\vecmap-\lone & 50.1 & 28.2 & \textbf{30.3} & \textbf{37.1} & \textbf{40.1} & 67.6 & 35.9\\   \midrule \midrule
\multicolumn{7}{l}{\textit{Supervised}}\\ 
\midrule
DLV\textsuperscript{\ETX} & 42.0 & 16.7 & 18.4 & 24.4 & 26.4 & 58.5 & 20.9 \\    
RCSLS\textsuperscript{\ETX} & 45.3 & 20.1 & 21.4 & 27.2 & 29.1 & 63.7 & 24.6 \\ \midrule
\joint-RSCLS & 46.6 & 20.9 & 22.1 & 29.0 & 29.9 & 65.2 & 25.3 \\
\joint-RSCLS-\ltwo & 46.4 & 20.8 & 22.3 & 29.0 & 29.8 & 65.2 & 25.3 \\ \hdashline
\joint-RSCLS-\lone & 47.3 & 22.2 & 23.8 & 30.1 & 31.2 & 65.9 & 26.6 \\ \midrule
\procb\textsuperscript{\ETX} & 50.7 & 25.0 & 26.3 & 32.8 & 34.8 & 66.5 & 29.8 \\ 
\procb-\ltwo & 50.0 & 24.1 & 25.6 & 31.8 & 34.3 & 66.4 & 29.6 \\ \hdashline
\procb-\lone & \textbf{51.1} & \textbf{25.6} & \textbf{26.9} & \textbf{33.6} & \textbf{35.0} & \textbf{67.4} & \textbf{30.5} \\  
                           \bottomrule
\end{tabular}
\caption{MRR (\%) of BLI for non-\en language pairs. 
Rows marked with \ETX are from the supplementary of \citet{xling}.
\muse yielded one unsuccessful run for \tr--\ita, and we only record the average of the two successful scores with *.
}
\label{tab:non_en}
}
\hfill \setcounter{table}{0} % numbering as figure
\parbox{.40\linewidth}{
\centering
    \centering
    \begin{subfigure}[t]{0.4\textwidth}
         \centering
         \includegraphics[width=\textwidth, trim={7.5cm 2.8cm 7.5cm 3cm},clip]{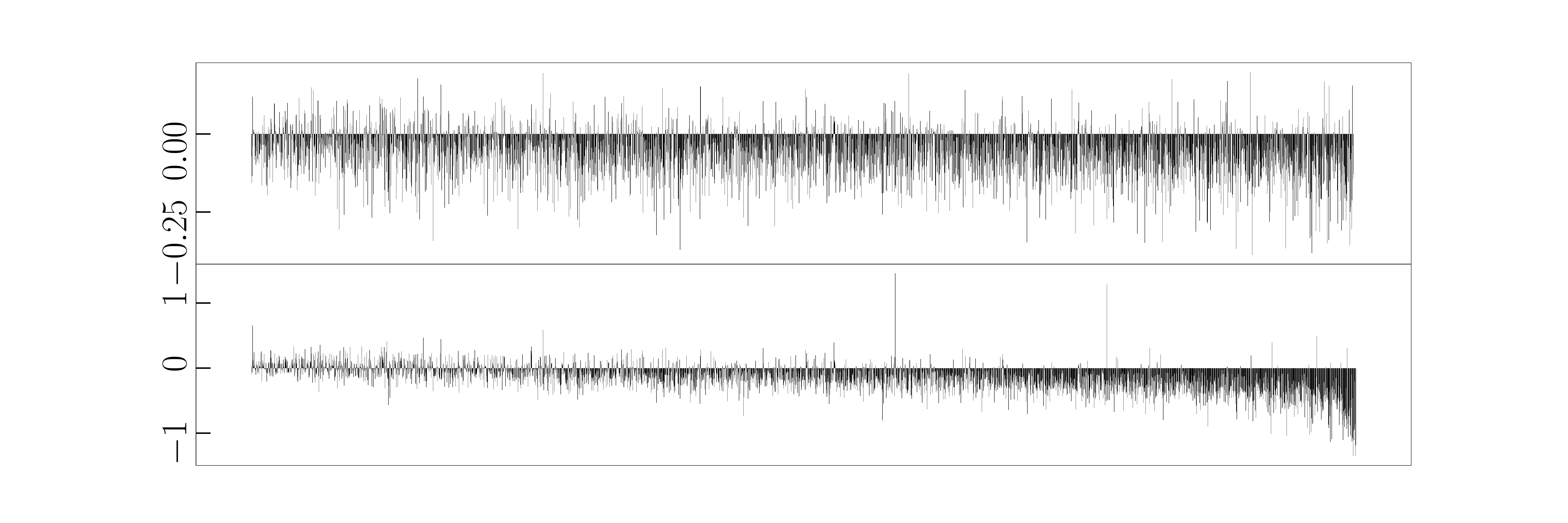}
         \caption{\vecmap on \en-\ru Wiki-Embs (cf. Tab.~\ref{tab:muse_dataset}).}
         \label{subfig:muse}
     \end{subfigure}
     ~
     \begin{subfigure}[t]{0.4\textwidth}
         \centering
         \includegraphics[width=\textwidth, trim={7.5cm 2.8cm 7.5cm 3cm},clip]{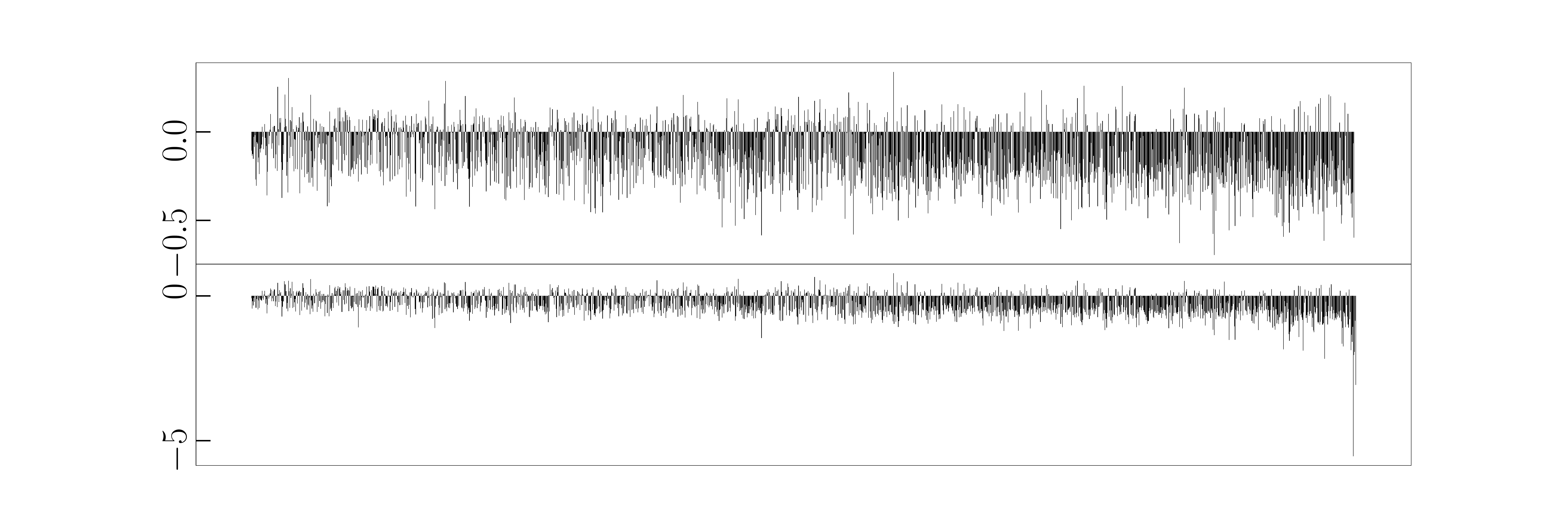}
         \caption{\procb on \en-\fin News-Embs (cf. Tab.~\ref{tab:procb_news}).}
         \label{subfig:vecmap}
     \end{subfigure}
     ~
    \begin{subfigure}[t]{0.4\textwidth}
         \centering
         \includegraphics[width=\textwidth, trim={7.5cm 2.8cm 7.5cm 3cm},clip]{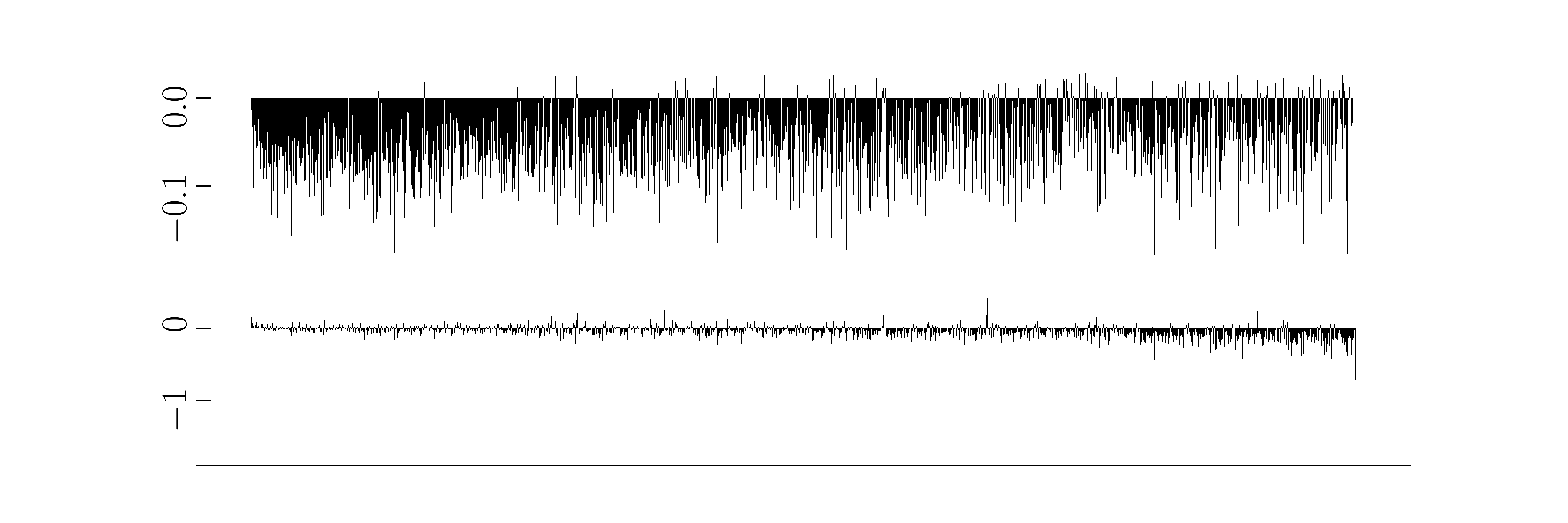}
         \caption{\muse on \ita-\fr Wiki-Embs (cf. Tab.~\ref{tab:non_en}).}
         \label{subfig:procb}
     \end{subfigure}
\captionof{figure}{Changes to $||\mathbf{A}\mathbf{M}-\mathbf{B}||_2$ after applying \lone\ (upper) and \ltwo\ (lower) refinement. Each word pairs is represented by a bar ordered on the x-axis by the distance between them.
See Appendix~\ref{sec:app} for alternative version.}
\label{fig:gap}
}
\end{table*}

\paragraph{Refining supervised baselines.}
To test the generalisability of our method, we also applied it on state-of-the-art supervised CLWE models: \procb~\cite{xling} and \joint-RCSLS~\cite{joint}. 
Following the setup of \citet{xling}, we learn mappings using Wiki-Embs and 1K training splits of their dataset.

Their evaluation code retrieves bilingual word pairs using the classic nearest-neighbour algorithm and outputs the Mean Reciprocal Rank (MRR).
As shown in Tab.~\ref{tab:procb_wiki}, both \joint-RCSLS-\lone\ and \procb-\lone\ outperform the baseline algorithms for all language pairs (with the exception of \en--\ita where the score of \procb is unchanged) with an average improvement of 0.9\% and 0.5\%, respectively ($p$\textless0.01). 

\joint-RCSLS-\lone\ and \procb-\lone\ were also tested using News-Embs with results shown in Tab.~\ref{tab:procb_news}\footnote{Note that results for \en--\es is not included, as no \en--\es dictionary is provided in \citet{xling}'s dataset.}. 
\lone\ refinement achieves an impressive improvement for both close (\en--$\{\de, \ita\}$) and distant (\en--\fin) language pairs: average gain of 1.9\% and 3.9\% respectively and over 5\% for \en--\de (\procb-\lone) in particular.
The \ltwo\ refinement does not benefit the supervised baseline, similar to the lack of improvement observed in the unsupervised setups.

\paragraph{Comparison of unsupervised and supervised settings.}
This part provides a comparison of the effectiveness of \lone\ refinement in unsupervised and supervised scenarios. 
Unlike previous experiments where only alignments involving English were investigated, these tests focus on non-\en setups.
\citet{xling}'s dataset is used to construct seven representative pairs which cover every category of etymological combination, i.e., intra-language-branch $\{\hr$--\ru, \ita--$\fr\}$, inter-language-branch $\{\de$--$\ita\}$, and  inter-language-family $\{\de$--\tr, \fin--\hr, \fin--\ita, \tr--$\ita\}$. The 1K training splits are used as seed lexicons in supervised runs. 
Apart from our main baselines, we further report the results of several other competitive CLWE models: 
Iterative Closest Point Model~\cite[ICP,][]{icp}, Gromov-Wasserstein Alignment Model~\cite[GWA,][]{gwa}, Discriminative Latent-Variable Model~\cite[DLV,][]{dlv} and 
Relaxed CSLS Model~\cite[RCSLS,][]{rcsls}.

Results shown in Tab.~\ref{tab:non_en} demonstrate that the main baselines (\muse, \joint-\muse, \vecmap, \joint-RCSLS, and \procb) outperform these other models by a large margin.
For all these main baselines, post applying \lone\ refinement improves the mapping quality for all language pairs ($p < 0.01$), with average improvements of 1.7\%, 1.4\%, 1.8\%, 1.1\%, and 0.8\%, respectively. 
Consistent with findings in the previous experiments, \ltwo\ refinement does not enhance performance. Improvement with \lone\ refinement is higher when language pairs are more distant, e.g., for all inter-language-family pairs such as \fin--\hr and \tr--\ita, even the minimum improvement of \muse-\lone\ over \muse is 2.3\%.

Comparing unsupervised and supervised approaches, it can be observed that \muse, \joint-\muse and \vecmap achieve higher overall gain with \lone\ refinement than \joint-RCSLS and \procb, where \joint-\muse-\lone\ and \vecmap-\lone\ give the best overall performance. 
One possible explanation to this phenomenon is that there is only a single source of possible noise in unsupervised models (i.e. the embedding topology) but for supervised methods noise can also be introduced via the seed lexicons.
Consequently unsupervised approaches drive more benefit from \lone\ refinement, which reduces the influence of topological outliers in CLWEs.

\paragraph{Topological behaviours of \lone\ and \ltwo\ refinement.}
To validate our assumption that \ltwo\ refinement is more sensitive to outliers while its \lone\ counterpart is more robust, 
we analyse how each refinement strategy changes the distance between bilingual word vector pairs in the synthetic dictionary $D$ (cf. Algorithm~\ref{algo:l1}) constructed from trained CLWE models. Specifically, for each word vector pair we subtract its post-refinement distance from the original distance (i.e., without applying additional \lone\ or \ltwo\ refinement step).  
Fig.~\ref{fig:gap} shows visualisation examples for three algorithms and language pairs, where each bar represents one word pair. 
It can be observed that \lone\ refinement effectively reduces the distance for most word pairs, regardless of their original distance (i.e., indicated by bars with negative values in the figures). The conventional \ltwo\ refinement strategy, in contrast, exhibits very different behaviour and tends to be overly influenced by word pairs with large distance (i.e. by outliers). The reason for this is that the \ltwo-norm penalty increases quadratically, causing the solution to put much more weight on optimising distant word pairs (i.e., word pairs on the right end of the X-axis show sharp distance decrements). This observation is in line with \citet{l2_bad} and  explains why \lone\ loss performs substantially stronger than \ltwo\ loss in the refinement.

\paragraph{Case study.} After aligning \en-\ru embeddings with unsupervised \muse, we measured the distance between vectors corresponding to the ground-truth dictionary of \citet{muse} (cf. Fig.~\ref{subfig:muse}). We then detected large outliers by finding vector pairs whose distance falls above $\mathrm{Q3}+1.5\cdot(\mathrm{Q3}-\mathrm{Q1})$, where $\mathrm{Q1}$ and $\mathrm{Q3}$ respectively denote the lower and upper quartile  based on the popular Inter-Quartile Range~\cite{Interquartile}. We found that many of the outliers correspond to polysemous entries, such as \{state (2$\times$ noun meanings and 1$\times$ verb meaning), \foreignlanguage{russian}{состояние} (only means \textit{status})\}, \{type (2$\times$ nominal meanings and 1$\times$ verb meaning), \foreignlanguage{russian}{тип} (only means \textit{kind})\}, and \{film (5$\times$ noun meanings), \foreignlanguage{russian}{фильм} (only means \textit{movie})\}.
We then re-perform \ltwo-based mapping after removing these vector pairs, observing that the accuracy jumps to 45.9\% (cf. the original \ltwo-norm alignment it is 43.8\% and after \lone\ refinement it is 45.6\%, cf. Tab.~\ref{tab:unsup_en}). This indicates that although all baselines already make use of preprocessing steps including vector normalization, outlier issues still exist and harms the \ltwo\ norm CLWEs. However, they can be alleviated by the proposed \lone\ refinement technique.

\subsection{Natural Language Inference}\label{subsec:NLI}

\setcounter{table}{3} % new numbering
\begin{table}[!t]\small
\centering
\begin{tabular}{{p{2.2cm}P{1.3cm}P{1.3cm}P{1.3cm}P{1.3cm}}}
\toprule 
\textit{Unsupervised}& \en--\de & \en--\fr & \en--\ru & \en--\tr \\ \midrule
ICP\textsuperscript{\ETX} & 58.0 & 51.0 & 57.2 & 40.0 \\  
GWA\textsuperscript{\ETX} & 42.7 & 38.3 & 37.6 & 35.9 \\  \midrule
\muse\textsuperscript{\ETX} & 61.1 & 53.6 & 36.3 & 35.9 \\  
\muse-\ltwo & 61.1 & 53.0 & \textcolor{white}{*}57.3* & \textcolor{white}{*}48.9* \\ \hdashline
\muse-\lone & \textbf{63.5} & 55.3 & \textcolor{white}{*}58.9* & \textcolor{white}{*}52.3* \\ \midrule
\joint-\muse & 61.3 & 55.2 & 58.1 & 55.0 \\
\joint-\muse-\ltwo & 61.2 & 55.2 & 57.6 & 55.1 \\ \hdashline
\joint-\muse-\lone & 62.9 & 57.9 & 59.4 & \textbf{57.5} \\ \midrule
\vecmap\textsuperscript{\ETX} & 60.4 & 61.3 & 58.1 & 53.4 \\ 
\vecmap-\ltwo & 60.3 & 60.6 & 57.7 & 53.5 \\ \hdashline
\vecmap-\lone & 61.5 & \textbf{63.7} & \textbf{60.1} & 56.4 \\  \midrule \midrule
\multicolumn{5}{l}{\textit{Supervised}}\\ 
\midrule
RCSLS\textsuperscript{\ETX} & 37.6 & 35.7 & 37.8 & 38.7 \\ \midrule
\joint-RSCLS & 50.2 & 48.9 & 51.0 & 51.7 \\
\joint-RSCLS-\ltwo & 50.4 & 48.6 & 50.9 & 51.5 \\ \hdashline
\joint-RSCLS-\lone & 51.3 & 50.1 & 53.2 & 52.6 \\ \midrule
\procb\textsuperscript{\ETX} & 61.3 & 54.3 & 59.3 & 56.8 \\
\procb-\ltwo & 61.0 & \textbf{54.8} & 58.9 & 55.1 \\ \hdashline
\procb-\lone & \textbf{62.1} & \textbf{54.8} & \textbf{60.7} & \textbf{58.2} \\  
                           \bottomrule
\end{tabular}
\caption{ACC (\%) of NLI. 
Rows marked with \ETX are from \citet{xling}.
\muse yielded one unsuccessful run for \en--\ru and \en--\tr respectively, which we exclude when calculating the average (with *).}
\label{tab:nli}
\end{table}
Finally, we experimented with a downstream NLI task in which the aim is to determine whether a ``hypothesis'' is true (\textit{entailment}), false (\textit{contradiction}) or undetermined (\textit{neutral}), given a ``premise''.
Higher ACC indicates better encoding of semantics in the tested embeddings.
The CLWEs used are those trained with Wiki-Embs for BLI. For \muse, \joint-\muse and \vecmap, we also obtain CLWEs for \en--\tr pair with the same configuration. 

Following  \citet{xling}, we first train the Enhanced Sequential Inference Model~\citep{esim} based on the large-scale English MultiNLI corpus~\citep{multinli} using vectors of language $\mathtt{L_A}$ (\en)  from an aligned bilingual embedding space (e.g., \en--\de). Next, we replace the $\mathtt{L_A}$ vectors with the vectors of language $\mathtt{L_B}$ (e.g., \de), and directly test the trained model on the language $\mathtt{L_B}$ portion of the XNLI corpus~\citep{xnli}.

Results in Tab.~\ref{tab:nli} show that the CLWEs refined by our algorithm yield the highest ACC for all language pairs in both supervised and unsupervised settings. The \ltwo\ refinement, on the contrary, is not beneficial overall. Improvements in cross-lingual transfer for NLI exhibit similar trends to those in the BLI experiments, i.e. greater performance gain for unsupervised methods and more distant language pairs, consistent with previous observations \cite{xling}. 
For instance, \muse-\lone\, \joint-\muse-\lone\ and \vecmap-\lone\  outperform their baselines by at least 2\% in ACC on average ($p < 0.01$), whereas the  improvements of \joint-RSCLS-\lone\ and \procb-\lone\ over their corresponding base methods are 2\% and 2.1\% respectively ($p < 0.01$). 
For both unsupervised and supervised methods, \lone\ refinement demonstrates stronger effect for more distant language pairs, e.g., \muse-\lone\  surpasses \muse by 1.2\% for \en--\fr, whereas a more impressive 2.7\% gain is achieved for \en--\tr.

In summary, in addition to improving BLI performance, our \lone\ refinement method also produces a significant improvement for a downsteam task (NLI),  demonstrating its effectiveness in improving the CLWE quality.

\section{Conclusion and Future Work}
This paper proposes a generic post-processing technique to enhance CLWE performance based on optimising \lone\ loss. 
This algorithm is motivated by successful applications in other research fields (e.g. computer vision and data mining) which exploit the \lone\ norm cost function since it has been shown to be more robust to noisy data than  the commonly-adopted \ltwo\ loss. The approach was evaluated using ten diverse languages and word embeddings from different domains on the popular BLI benchmark, as well as a downstream task of cross-lingual transfer for NLI. Results demonstrated that our algorithm can significantly improve the quality of CLWEs in both supervised and unsupervised setups. It is therefore recommended that this straightforward technique be applied to improve performance of CLWEs.

The convergence speed of the optimiser prevented us from performing \lone\ loss optimisation over multiple iterations.
Future work will focus on improving the efficiency of our \lone\ OPA solver, as well as exploring the application of other robust loss functions within CLWE training strategies.

\section*{Ethics Statement}
This work provides an effective post-hoc method to improve CLWEs, advancing the state-of-the-art in both supervised and unsupervised settings. Our comprehensive empirical studies demonstrate that the proposed algorithm can facilitate researches in machine translation, cross-lingual transfer learning, etc, which have deep societal impact of bridging cultural gaps across the world.

Besides, this paper introduces and solves an optimisation problem based on an under-explored robust cost function, namely \lone\ loss. We believe it could be of interest for the wider community as outlier is a long-standing issue in many artificial intelligence applications.

One caveat with our method, as is the case for all word-embedding-based systems, is that various biases may exist in vector spaces. We suggest this problem should always be looked at critically. In addition, our implemented solver can be computationally expensive, leading to increased electricity consumption and the associated negative environmental repercussions.

\section*{Acknowledgements}
This work is supported by the award made by the UK Engineering and Physical Sciences Research Council (Grant number: EP/P011829/1) and Baidu,~Inc.
We would also like to express our sincerest gratitude to Guanyi Chen, Ruizhe Li, Xiao Li, Shun Wang, and the anonymous reviewers for their insightful and helpful comments.

\bibliography{l1opa}
\bibliographystyle{acl_natbib}
\clearpage
\appendix

\newgeometry{left=2cm,top=0cm,right=2cm,bottom=1cm}
\renewcommand\thefigure{\thesection.\arabic{figure}}
\setcounter{figure}{0}

\newcommand\invisiblesection[1]{%
  \refstepcounter{section}%
  \addcontentsline{toc}{section}{\protect\numberline{\thesection}#1}%
  \sectionmark{#1}}

\invisiblesection{}~\label{sec:app}
\begin{figure*}[h!]
        \centering
    \begin{subfigure}[t]{0.75\textwidth}
         \centering
         \includegraphics[width=\textwidth]{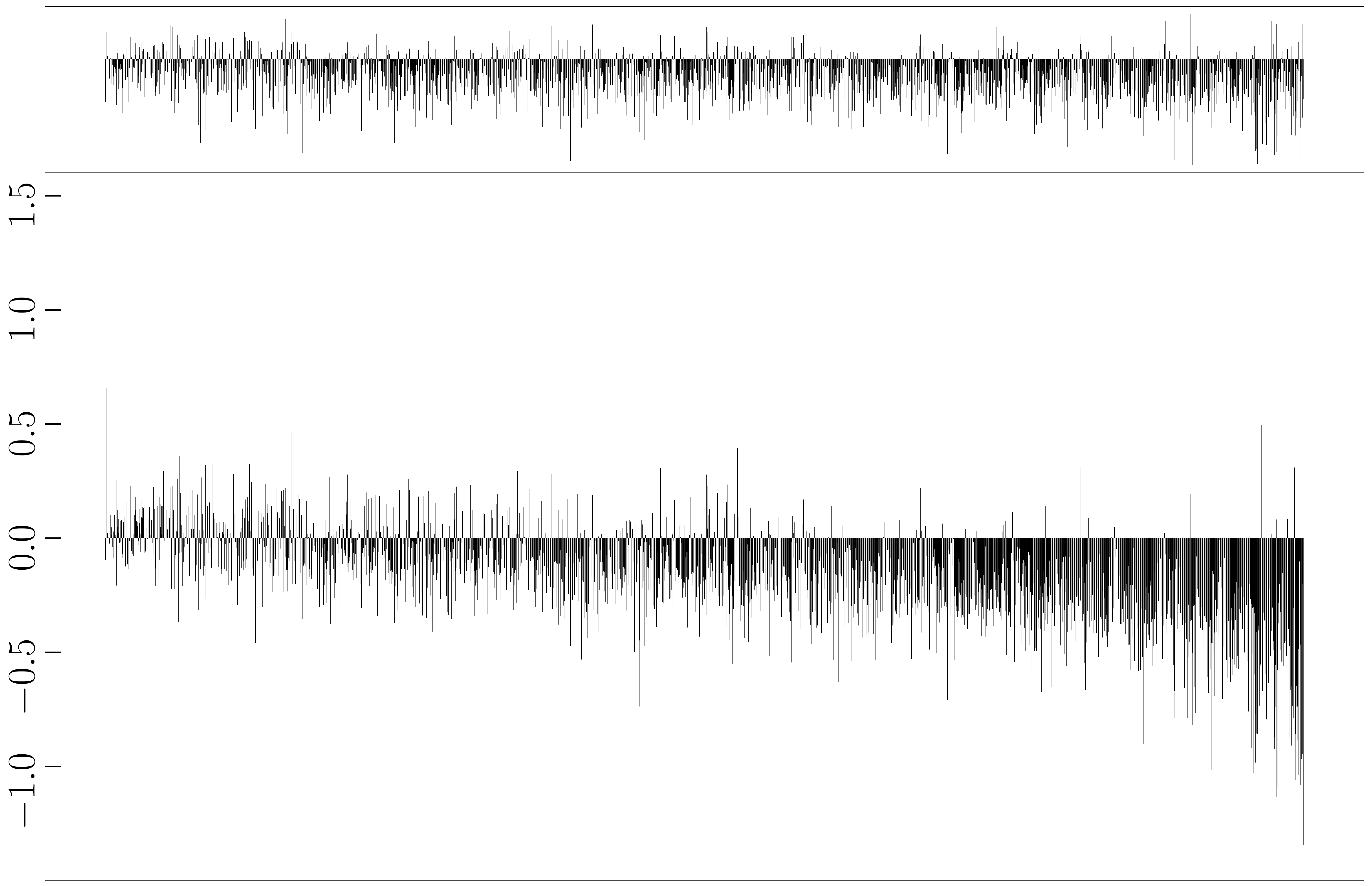}
         \caption{\vecmap on \en-\ru Wiki-Embs (cf. Tab.~\ref{tab:muse_dataset}).}
     \end{subfigure}
     ~
     \begin{subfigure}[t]{0.75\textwidth}
         \centering
         \includegraphics[width=\textwidth]{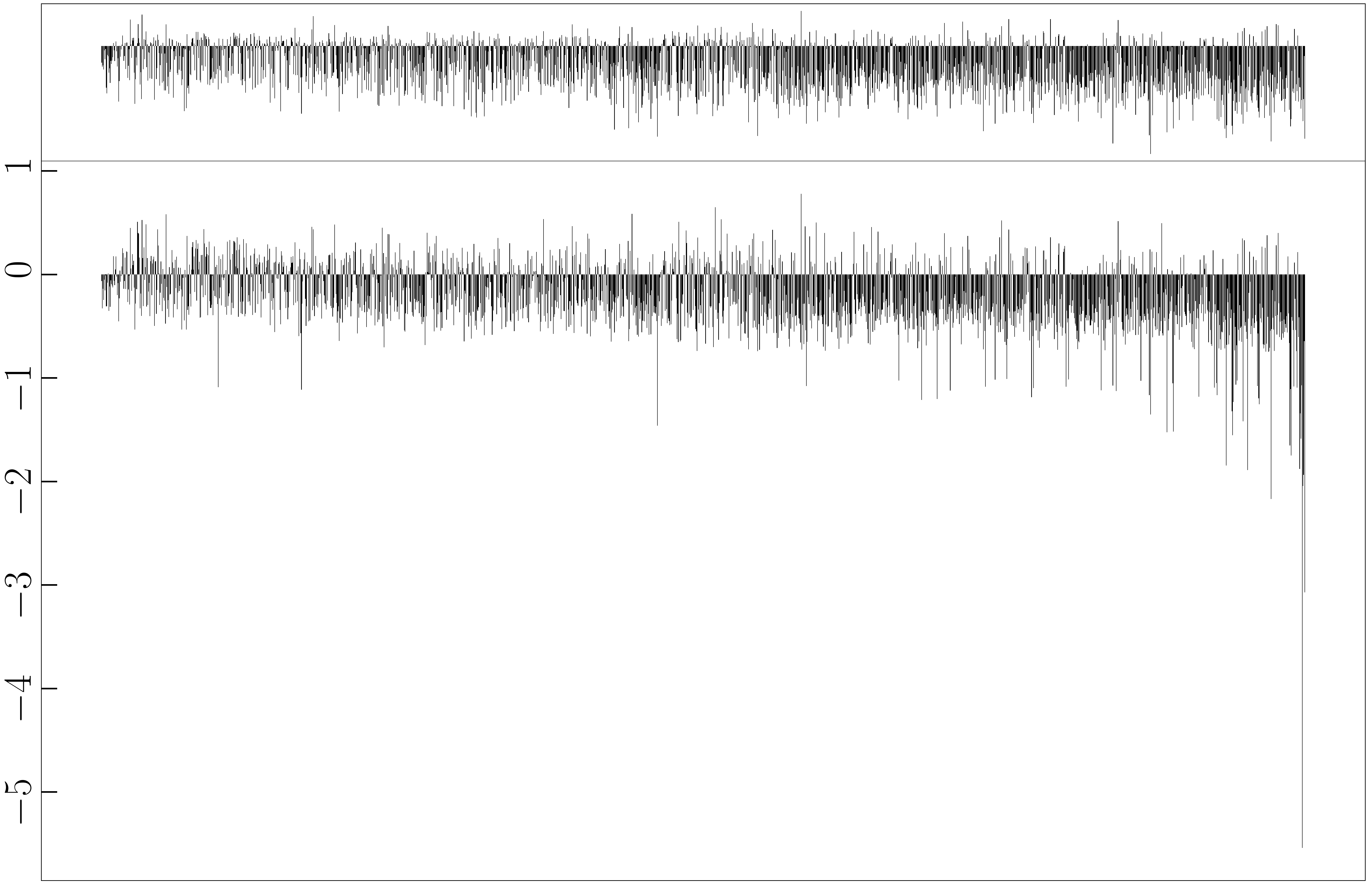}
         \caption{\procb on \en-\fin News-Embs (cf. Tab.~\ref{tab:procb_news}).}
     \end{subfigure}
     ~
    \begin{subfigure}[t]{0.75\textwidth}
         \centering
         \includegraphics[width=\textwidth]{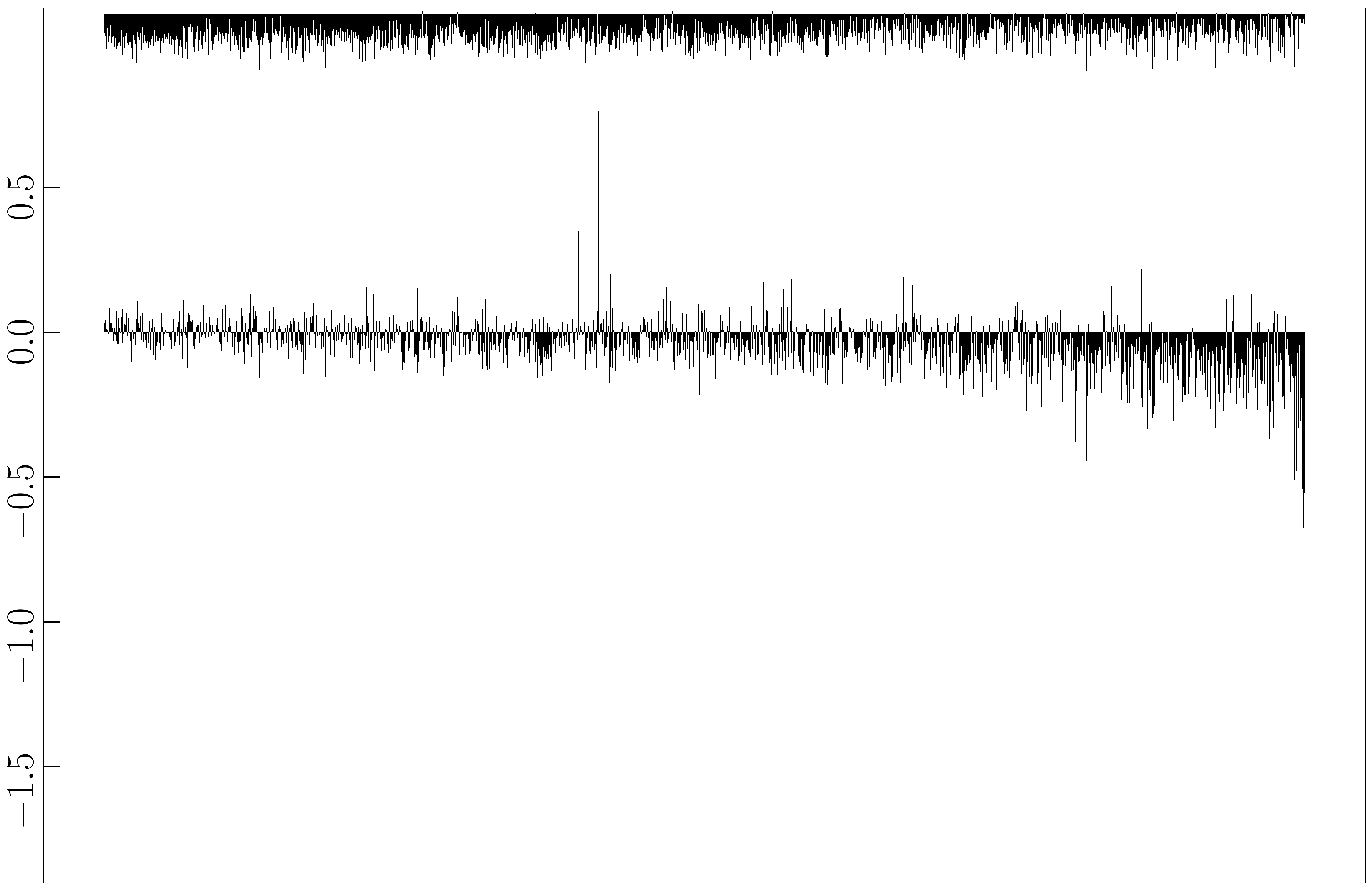}
         \caption{\muse on \ita-\fr Wiki-Embs (cf. Tab.~\ref{tab:non_en}).}
     \end{subfigure}
    \caption{Changes to $||\mathbf{A}\mathbf{M}-\mathbf{B}||_2$ after applying \lone\ (upper) and \ltwo\ (lower) refinement. Different from Fig.~\ref{fig:gap}, in each sub-figure the upper and lower Y-axis scales are uniformed.}
    \label{fig:app}
\end{figure*}

\end{document}